\documentclass[11pt,fullpage, letterpaper,twoside]{article}

\usepackage[margin=1in]{geometry}

\usepackage[utf8]{inputenc} 
\usepackage[T1]{fontenc}    
\usepackage{hyperref}       
\usepackage{url}            
\usepackage{booktabs}       
\usepackage{amsfonts}       
\usepackage{nicefrac}       
\usepackage{microtype}      
\usepackage{algorithm,algpseudocode}
\usepackage{mathtools}
\usepackage{amsmath,amssymb}
\usepackage{graphicx}
\usepackage{subfigure}
\usepackage{color}
\DeclareMathOperator{\E}{\mathbb{E}}
\title{Task-Agnostic Meta-Learning for Few-shot Learning}

\author{Muhammad Abdullah Jamal$^\dag$, Guo-Jun Qi$^\dag$\footnote{Corresponding author: G.-J. Qi, email: guojunq@gmail.com.}, and Mubarak Shah$^\sharp$ \\\\
$\dag$ Laboratory for {\bf MA}chine {\bf P}erception and {\bf LE}arning \\
\url{http://maple.cs.ucf.edu}\\
$\sharp$ Center for Research in Computer Vision\\
University of Central Florida\\
}

\date{}

\begin{document}

\maketitle

\begin{abstract}
Meta-learning approaches have been proposed to tackle the few-shot learning problem. 
Typically, a meta-learner is trained on a variety of tasks in the hopes of being generalizable to new tasks. However, the generalizability on new tasks of a meta-learner could be fragile  when it is over-trained on existing tasks during meta-training phase. In other words, the initial model of a meta-learner could be too biased towards existing tasks to adapt to new tasks, especially when only very few examples are available to update the model. To avoid a biased meta-learner and improve its generalizability, we propose a novel paradigm of Task-Agnostic Meta-Learning (TAML) algorithms. Specifically, we present an entropy-based approach that meta-learns an unbiased initial model with the largest uncertainty over the output labels by preventing it from over-performing in classification tasks. Alternatively, a more general inequality-minimization TAML is presented for more ubiquitous scenarios by directly minimizing the inequality of initial losses beyond the classification tasks wherever a suitable loss can be defined.
Experiments on benchmarked datasets demonstrate that the proposed approaches outperform compared meta-learning algorithms in both few-shot classification and reinforcement learning tasks.
\end{abstract}

\section{Introduction}
The key to achieving human level intelligence is to learn from a few labeled examples. Human can learn and adapt quickly from a few examples using prior experience. We want our learner to be able to learn from a few examples and quickly adapt to a changing task. All these concerns motivate to study the few-shot learning problem. The advantage of studying the few-shot problem is that it only relies on few examples and it alleviates the need to collect large amount of labeled training set which is a cumbersome process.

Recently, meta-learning approach is being used to tackle the problem of few-shot learning. A meta-learning model usually contains two parts  -- an initial model, and an updating strategy (e.g., a parameterized model) to train the initial model to a new task with few examples. 
Then the goal of meta-learning is to automatically
meta-learn the optimal parameters for both the initial model and the updating strategy that are generalizable across a variety of tasks.  There are many meta-learning approaches that show promising results on few-shot learning problems.
For example, Meta-LSTM~\cite{Sachin2017} uses LSTM meta-learner that not only learns initial model but also the updating rule. On the contrary, MAML~\cite{DBLP:journals/corr/FinnAL17} only learns an initial model since its updating rule is fixed to a classic gradient descent method as a meta-learner.

The problem with existing meta-learning approaches is that the initial model can be trained biased towards some tasks, particularly those sampled in meta-training phase. Such a biased initial model may not be well generalizable to an unseen task that has a large deviation from meta-training tasks, especially when very few examples are available on the new task.  This inspires us to meta-train an unbiased initial model by preventing it from overperforming on some tasks or directly minimizing the inequality of performances across different tasks, in a hope to make it more generalizable to unseen tasks.
To this end, we propose a Task-Agnostic Meta-Learning (TAML) algorithms in this paper.




Specifically, we propose two novel paradigms of TAML algorithms -- an entropy-based TAML and inequality-minimization measures based TAML. The idea of using entropy based approach is to maximize
the entropy of labels predicted by the initial model to prevent it from overperforming on some tasks.  However, the entropy-based approach is limited to discrete outputs from a model, making it more amenable to classification tasks.

The second paradigm is inspired by inequality measures used in Economics. The idea is to meta-train an initial model in such a way that it directly minimizes the inequality of losses by the initial model across a variety of tasks. This will force the meta-learner to learn a unbiased initial model without over-performing on some particular tasks. Meanwhile, any form of losses can be adopted for involved task without having to rely on discrete outputs. This makes this paradigm more ubiquitous to many scenarios beyond classification tasks.



The remainder of the paper is organized as follows. We elaborate the proposed TAML approach to meta-learning in Section~\ref{sApproach}. It is followed by a review  about the related work (Section~\ref{sRelated}). In Section~\ref{sExp}, we present extensive experimental studies on few-shot classification and reinforcement learning.

\vspace{-5pt}
\section{Approach}\label{sApproach}
Our goal is to train a model that can be task-agnostic in a way that it prevents the initial model or learner to over-perform on a particular task. In this section, we will first describe our entropy based and inequality-minimization measures based approach to the problem, and then we will discuss some of the inequality measures that we used in the paper.

\subsection{Task Agnostic Meta-Learning}\label{TAML}
In this section, we propose a task-agnostic approach for few-shot meta-learning. 
The goal of few-shot meta-learning is to train a model in such a way that it can learn to adapt rapidly using few samples for a new task. In this meta-learning approach, a learner is trained during a meta-learning phase on variety of sampled tasks so that it can learn new tasks 
, while a meta-learner trains the learner and is responsible for learning the update rule and initial model.

The problem with the current meta-learning approach is that the initial model or learner can be biased towards some tasks sampled during the meta-training phase, particularly when future tasks in the test phase may have discrepancy from those in the training tasks. In this case, we wish to avoid an initial model over-performing on some tasks. Moreover, an over-performed initial model could also prevent the meta-learner to learn a better update rule with consistent performance across tasks.

To address this problem, we impose an unbiased task-agnostic prior on the initial model by preventing it from over-performing on some tasks so that a meta-learner can achieve a more competitive update rule. There have been many meta-learning approaches to few-shot learning problems that have been briefly discussed in the section~\ref{sRelated}. While the task-agnostic prior is a widely applicable principle for many meta-learning algorithms, we mainly choose Model-Agnostic Meta Learning approach (MAML) as an example to present the idea, and it is not hard to extend to other meta-learning approaches.

In the following, we will depict the idea by presenting two paradigms of task-agnostic meta-learning (TAML) algorithms -- the entropy-maximization/reduction TAML and inequality-minimization TAML.


\subsubsection{Entropy-Maximization/Reduction TAML}
For simplicity, we express the model as a function $f_\theta$ that is parameterized by $\theta$. For example, it can be a classifier that takes an input example and outputs its discrete label. During meta-training, a batch of tasks are sampled from a task distribution $p(\mathcal{T})$, and each task is $K$-shot $N$-way problem where $K$ represents the number of training examples while $N$ represent the number of classes depending on the problem setting. In the MAML, a model is trained on a task $\mathcal{T}{_i}$ using $K$ examples and then tested on a few new examples $\mathcal{D}_{val}$ for this task.

A model has an initial parameter $\theta$ and when it is trained on the task $\mathcal{T}{_i}$, its parameter is updated from $\theta$ to $\theta_i$ by following an updating rule. For example, for $K$-shot classification,
stochastic gradient descent can be used to update model parameter by
$
\theta_i \leftarrow \theta - \alpha \nabla_\theta \mathcal L_{\mathcal T_i}(f_\theta)
$
that attempts to minimize the cross-entropy loss $\mathcal L_{\mathcal T_i}(f_\theta)$ for the classification task $\mathcal{T}_i$ over $K$ examples.

To prevent  the initial model $f_\theta$ from over-performing on a task, we prefer it makes a random guess over predicted labels with an equal probability so that it is not biased towards the task. This can be expressed as a maximum-entropy prior over $\theta$ so that the initial model should have a large entropy over the predicted labels over samples from task $\mathcal{T}{_i}$.

The entropy for task $\mathcal T_i$ is computed by sampling  $x_i$ from $P_{\mathcal T_i} (x)$ over its output probabilities $y_{i,n}$ over $N$ predicted labels:
\begin{equation}
\mathcal{H}_{\mathcal T_i}(f_\theta) = -\mathbb E_{x_i\sim P_{\mathcal T_i} (x)} \sum_{n=1}^N \hat{y}_{i,n} \log(\hat{y}_{i,n})\label{entropy}
\end{equation}
where $[y_{i,1},\cdots,y_{i,N}]=f_\theta(x_i)$ is the predictions by $f_\theta$, which are often an output from a softmax layer in a classification task. The above expectation is taken over $x_i$'s sampled from task $\mathcal T_i$.

Alternatively, one can not only maximize the entropy before the update of initial model's parameter, but also minimize the entropy after the update. So overall, we maximize the entropy reduction
for each task $\mathcal{T}{_i}$ as
$\mathcal{H}_{\mathcal{T}_i}(f_{\theta}) - \mathcal{H}_{\mathcal{T}_i}(f_{\theta_i})$.
The minimization of $\mathcal{H}_{\mathcal{T}_i}(f_{\theta_i})$ means that the model can become more certain about the labels with a higher confidence after updating the parameter $\theta$ to $\theta_i$.
This entropy term can be combined with the typical meta-training objective term as a regularizer to find the optimal $\theta$, which is
$$
\min_\theta \mathbb E_{\mathcal T_i \sim P(\mathcal T)} \mathcal L_{\mathcal T_i} (f_{\theta_i}) + \lambda [-\mathcal{H}_{\mathcal{T}_i}(f_{\theta}) + \mathcal{H}_{\mathcal{T}_i}(f_{\theta_i})]
$$
where $\lambda$ is a positive balancing coefficient, and the first term is the expected loss for the updated model $f_{\theta_i}$. The entropy-reduction algorithm is summarized in ~\ref{entTaml}.

\begin{algorithm}
\begin{algorithmic}
\Require $p(\mathcal{T})$: distribution over tasks.
\Require $\alpha,\beta$: hyperparameters\\
\noindent Randomly Initialize $\theta$\
\While{not done}
   \State Sample batch of tasks $\mathcal{T}{_i} \sim  p(\mathcal{T})$
   \ForAll{$\mathcal{T}{_i}$}
     \State Sample $K$ samples from $\mathcal{T}{_i}$
     \State Evaluate $\nabla{_\theta} \mathcal{L}_{\mathcal{T}_i}(f{_\theta})$ and
     $\mathcal{L}_{\mathcal{T}_i}(f_\theta)$ using $K$ samples .
     \State Compute adapted parameters using gradient descent
     \State $\theta_i \leftarrow \theta - \alpha \nabla_{\theta} \mathcal{L}_{\mathcal{T}_i}$
     \State Sample $\mathcal{D}_{val}$ from $\mathcal{T}_{i}$ for meta update.
    \EndFor

    \State Update  $\theta \leftarrow \theta - \beta \nabla{_\theta} \{\mathbb E_{\mathcal T_i \sim P(\mathcal T)} \mathcal L_{\mathcal T_i} (f_{\theta_i}) + \lambda [-\mathcal{H}_{\mathcal{T}_i}(f_{\theta}) + \mathcal{H}_{\mathcal{T}_i}(f_{\theta_i})]\}$ using $\mathcal{D}_{val}$,   \State$\mathcal{L}_{\mathcal{T}_i}$, and $\mathcal{H}_{\mathcal{T}_i}$.

\EndWhile

\caption{Entropy-Reduction TAML for Few-Shot Classification}
\label{entTaml}
\end{algorithmic}
\end{algorithm}

Unfortunately, the entropy-based TAML is subject to a critical limitation -- it is only amenable to discrete labels in classification tasks to compute the entropy. In contrast, many other learning problems, such as regression and reinforcement learning problems, it is often trained by minimizing some loss or error functions directly without explicitly accessing a particular form of outputs like discrete labels. To make the TAML widely applicable, we need to define an alternative metric to measure and minimize the bias across tasks.

\vspace{-3pt}
\subsubsection{Inequality-Minimization TAML}
We wish to train a task-agnostic model in meta-learning such that its initial performance is  unbiased towards any particular task $\mathcal{T}{_i}$. Such a task-agnostic meta-learner would do so by minimizing the inequality of its performances over different tasks.

To this end, we propose an approach based on a large family of statistics used to measure the "economic inequalities" to measure the "task bias". The idea is that the loss of an initial model on each task $\mathcal{T}_i$ is viewed as an income for that task. Then for the TAML model, its loss inequality over multiple tasks is minimized to make the meta-learner task-agnostic.

Specifically, the bias of the initial model towards any particular tasks is minimized during meta-training by minimizing the inequality over the losses of sampled tasks in a batch. So, given an unseen task during testing phase, a better generalization performance is expected on the new task by updating from an unbiased initial model with few examples.
The key difference between both TAMLs lies that for entropy, we only consider one task at a time by computing the entropy of its output labels. Moreover, entropy  depends on a particular form or explanation of output function, e.g., the SoftMax output. On the contrary, the inequality only depends on the loss, thus it is more ubiquitous.

The complete algorithm is explained in ~\ref{ieTaml}. Formally, consider a batch of sampled tasks $\{\mathcal T_i\}$ and their losses $\{\mathcal L_{\mathcal T_i}(f_\theta)\}$ by the initial model $f_\theta$, one can compute the inequality measure by $\mathcal{I}_{\mathcal{E}}(\{\mathcal L_{\mathcal T_i}(f_\theta)\})$ as discussed later. Then the initial model parameter $\theta$ is meta-learned by minimizing the following objective
$$
\mathbb E_{\mathcal{T}_i  \sim  p(\mathcal{T})} \left[\mathcal{L}_{\mathcal{T}_i}(f_{\theta_i})\right] + \lambda\mathcal{I}_{\mathcal{E}}(\{\mathcal{L}_{\mathcal{T}_i}(f_{\theta})\})
$$
through gradient descent as shown in Algorithm~\ref{ieTaml}. It is worth noting that the inequality measure is computed over a set of losses from sampled tasks. The first term is the expected loss by the model $f_{\theta_i}$ after the update, while the second is the inequality of losses by the initial model $f_\theta$ before the update. Both terms are a function of the initial model parameter $\theta$ since $\theta_i$ is updated from $\theta$.
In the following, we will elaborate on some choices on inequality measures $\mathcal I_\mathcal E$.

\subsection{Inequality Measures}\label{sIeqMeas}
Inequality measures are instrumental towards calculating the economic inequalities in the outcomes that can be wealth, incomes, or health related metrics. In meta-learning context, we use $\ell_i=\mathcal L_{\mathcal T_i}(f_\theta)$ to represent the loss of a task $\mathcal{T}_i$, $\bar{\ell}$ represents the mean of the losses over sampled tasks, and $M$ is the number of tasks in a single batch. The inequality measures used in TAML are briefly described below.
\paragraph{Theil Index~\cite{theil1967economics}.}This inequality measure has been derived from redundancy in information theory, which is defined as the difference between the maximum entropy of the data and an observed entropy. 
Suppose that we have $M$ losses $\{\ell_i|i=1,\cdots,M\}$, then Thiel Index is defined as
\begin{equation}
T_T  = \frac{1}{M}\sum_{i=1}^M \frac{\ell_i}{\bar{\ell}}\ln\frac{\ell_i}{\bar{\ell}}\label{Thiel}
\end{equation}
\paragraph{Generalized Entropy Index~\cite{COWELL1980147}.}The relation between information theory and information distribution analysis has been exploited to derive a number of measures for inequality. Generalized Entropy index has been proposed to measure the income inequality. It is not a single inequality measure, but it is a family that includes many inequality measures like Thiel Index, Thiel L etc. For some real value $\alpha$, it is defined as:
\begin{equation}
    GE(\alpha)=
\begin{dcases}
   \frac{1}{M\alpha(\alpha-1)}\sum_{i=1}^M\bigg[\biggl(\frac{\ell_i}{\bar{\ell}}\biggr)^\alpha- 1\bigg] ,&  \alpha \neq 0,1,\\
     \frac{1}{M}\sum_{i=1}^M \frac{\ell_i}{\bar{\ell}}\ln\frac{\ell_i}{\bar{\ell}}, &\alpha = 1,\\
    -\frac{1}{M}\sum_{i=1}^M \ln\frac{\ell_i}{\bar{\ell}}, &\alpha = 0,
\end{dcases}\label{GE}
\end{equation}
From the equation, we can see that it does represent a family of inequality measures. When $\alpha$ is zero,  it is called a mean log deviation of Thiel L, and when $\alpha$ is one, it is actually Thiel Index. A larger GE $\alpha$ value makes this index more sensitive to differences at the upper part of the distribution, and a smaller $\alpha$ value makes it more sensitive to differences at the bottom of the distribution.
\paragraph{Atkinson Index~\cite{ATKINSON1970244}.}It is another measure for income inequality which is useful in determining which end of the distribution contributed the most to the observed inequality. It is defined as :
\begin{equation}
    A_\epsilon=
\begin{dcases}
    1 - \frac{1}{\mu}\biggl(\frac{1}{M}\sum_{i=1}^M \ell_i^{1-\epsilon}\biggr)^{\frac{1}{1-\epsilon}},&  \text{for } 0 \leq \epsilon \neq 1,\\
     1 - \frac{1}{\bar{\ell}}\biggl(\frac{1}{M}\prod_{i=1}^M \ell_i\biggr)^{\frac{1}{M}},&  \text{for } \epsilon = 1,,
\end{dcases}\label{AE}
\end{equation}
where $\epsilon$ is called "inequality aversion parameter". When $\epsilon = 0$ the index becomes more sensitive to the changes in upper end of the distribution ,and when it approaches to 1, the index becomes more sensitive to the changes in lower end of the distribution.
\paragraph{Gini-Coefficient~\cite{10.2307/2094626}.}It is usually defined as the half of the relative absolute mean difference. In terms of meta-learning, if there are M tasks in a single batch and a task $\mathcal{T}_i$ loss is represented by $\ell_i$, then Gini-Coefficient is defined as:
\begin{equation}
G = \frac{\sum_{i=1}^M\sum_{j=1}^M|\ell_i - \ell_j|}{2n\sum_{i=1}^M \ell_i}\label{gini}
\end{equation}
Gini- coefficient is more sensitive to deviation around the middle of the distribution than at the upper or lower part of the distribution.
\paragraph{Variance of Logarithms~\cite{RePEc:cvs:starer:97-22}.}It is another common inequality measure defined as:
\begin{equation}
V_L(\ell) = \frac{1}{M}\sum_{i=1}^M [\ln \ell_i - \ln g(\ell)]^2 \label{VL}
\end{equation}
where g($\ell$) is the geometric mean of $\ell$ which is defined as $(\prod_{i=1}^M\ell_i)^{1/M}$ . The geometric mean put greater emphasis on the lower losses of the distribution.

\begin{algorithm}
\begin{algorithmic}
\Require $p(\mathcal{T})$: distribution over tasks.
\Require $\alpha,\beta$: hyperparameters\\
\noindent Randomly Initialize $\theta$\
\While{not done}
   \State Sample batch of tasks $\mathcal{T}{_i} \sim  p(\mathcal{T})$
   \ForAll{$\mathcal{T}{_i}$}
     \State Sample $K$-shot samples from $\mathcal{T}{_i}$
     \State Evaluate $\nabla{_\theta} \mathcal{L}_{\mathcal{T}_i}(f{_\theta})$ and
     $\mathcal{L}_{\mathcal{T}_i}$ using $K$ samples .
     \State Compute adapted parameters using gradient descent
     \State $\theta_i = \theta - \alpha \nabla_{\theta} \mathcal{L}_{\mathcal{T}_i}$
     \State Sample a dataset $\mathcal{D}_{val,i}$ from task $\mathcal{T}_{i}$ used below.
    \EndFor
    \State Update  $\theta \leftarrow \theta - \beta \nabla_\theta[
    \mathbb E_{\mathcal{T}_i  \sim  p(\mathcal{T})} \mathcal{L}_{\mathcal{T}_i}(f_{\theta_i}) + \lambda\mathcal{I}_{\mathcal{E}}(\{\mathcal{L}_{\mathcal{T}_i}(f_{\theta})\})]$  using $\mathcal{D}_{val,i}$, $\mathcal{L}_{\mathcal{T}_i}$, and $\mathcal{I}_{\mathcal{E}}$

\EndWhile

\caption{Inequality Measures Based TAML for Few-Shot Classification}
\label{ieTaml}
\end{algorithmic}
\end{algorithm}

\section{Related Work}\label{sRelated}
The idea of meta-learning has been proposed more than a couple of decades ago ~\cite{schmidhuber:1987:srl,287172,Thrun:1998:LL:296635}. Most of the approaches to meta-learning include learning a learner's model by training a meta-learner. Recent studies towards meta-learning for deep neural networks include learning a hand-designed optimizer like SGD by parameterizing it through recurrent neural networks. Li~\cite{DBLP:journals/corr/LiM16b}, and Andrychowicz~\cite{DBLP:journals/corr/AndrychowiczDGH16} studied a LSTM based meta-learner that takes the gradients from learner and performs an optimization step.  Recently, meta-learning framework has been used to solve few-shot classification problems.~\cite{Sachin2017} used the same LSTM based meta-learner approach in which LSTM meta-learner takes the gradient of a learner and proposed an update to the learner's parameters. The approach learns both weight initialization and an optimizer of the model weights. Finn~\cite{DBLP:journals/corr/FinnAL17} proposed a more general approach for meta-learning known as MAML by simply learning weight initialization for a learner through a fixed gradient descent. It trains a model on a variety of tasks to have a good initialization point that can be quickly adapted (few or one gradient steps) to a new task using few training examples. Meta-SGD~\cite{DBLP:journals/corr/LiZCL17} extends the MAML, which not only learns weight initialization but also the learner's update step size. ~\cite{mishra2018a} proposes a temporal convolution and attention based meta-learner called SNAIL that achieves state-of-the-art performance for few-shot classification tasks and reinforcement learning tasks.

Other paradigms of meta-learning approaches include training a memory augmented neural network on existing tasks by coupling with LSTM or feed-forward neural network controller~\cite{DBLP:journals/corr/SantoroBBWL16,DBLP:journals/corr/MunkhdalaiY17}. There are also several non-meta-learning approaches to few-shot classification problem by designing specific neural architectures. For example, ~\cite{Koch2015SiameseNN} trains a Siamese network to compare new examples with existing ones in a learned metric space. Vinyals~\cite{DBLP:journals/corr/VinyalsBLKW16} used a differentiable nearest neighbour loss by utilizing the cosine similarities between the features produced by a convolutional neural network.~\cite{snell2017prototypical} proposed a similar approach to matching net but used a square euclidean distance metric instead. In this paper, we mainly focus on the meta-learning approaches and their applications to few-shot classiciation and reinforcement tasks.

\begin{table}[t!]
\centering
\caption{Few Shot Classification results on Omniglot dataset for fully connected network and convolutional network on 5-way setting, where * means re-run results as there is no general training/test splitting available for Omniglot, thus we re-run compared models with the same splitting used in running the TAML for a fair comparison.
The $\pm$ shows 95\% confidence interval over tasks.}
\label{table:omni5way}
\resizebox{0.79\columnwidth}{!}{%
\begin{tabular}{|l|l|l|l}
\cline{1-3}
Methods                                   & \multicolumn{2}{l|}{\begin{tabular}[c]{@{}l@{}}~ ~ ~ ~ ~ ~ ~ ~ ~ ~ ~ ~5-way\\~ ~1-shot ~ ~ ~ ~ ~ ~ ~ ~ ~ ~ ~ ~ ~5-shot~~\end{tabular}}                                               &   \\
\cline{1-3}
MANN, no conv~\cite{DBLP:journals/corr/SantoroBBWL16}                             & 82.8\%                                                                                        & 94.9\%                                      &   \\
\cline{1-3}
MAML, no conv~\cite{DBLP:journals/corr/FinnAL17}                             & 89.7 $\pm$ 1.1\%                                                                                  & 97.5 $\pm$ 0.6 \%(96.1 $\pm$ 0.4)\%*                &   \\
\cline{1-3}
\textbf{TAML(Entropy), no conv}           & \textbf{91.19 $\pm$ 1.03\%}                                                                        & \textbf{97.40 $\pm$ 0.34\%} &   \\
\cline{1-3}
\textbf{TAML(Theil), no conv}             & \textbf{91.37 $\pm$ 0.97\%}                                                                        & 96.84 $\pm$ 0.36\%                              &   \\
\cline{1-3}
\textbf{TAML(GE(2)), no conv}             & \textbf{91.3 $\pm$ 1.0\%}                                                                         & 96.76\textit{ }$\pm$ 0.4\%                      &   \\
\cline{1-3}
\textbf{TAML(Atkinson), no conv}          & \textbf{91.77 $\pm$ 0.97\%}                                                                        & 97.0 $\pm$ 0.4\%                                &   \\
\cline{1-3}
\textbf{TAML (Gini-Coefficient), no conv} & \textbf{93.17 $\pm$ 1.0\%}                                                                         & {-}                       & \\
\cline{1-3}
\cline{1-3}
Siamese Nets~\cite{Koch2015SiameseNN}                     & 97.3\%                                                                                                                                     & 98.4\%~                                                           &   \\
\cline{1-3}
Matching Nets~\cite{DBLP:journals/corr/VinyalsBLKW16}                    & 98.1\%                                                                                                                                     & 98.9\%~                                                           &   \\
\cline{1-3}
Neural Statistician~\cite{53ff0dc9643843d8ac4b40b699cb6bd8}            & 98.1\%                                                                                                                                     & 99.5\%~                                                           &   \\
\cline{1-3}
Memory Mod.~\cite{DBLP:journals/corr/KaiserNRB17}                      & 98.4\%                                                                                                                                     & 99.6\%~                                                           &   \\
\cline{1-3}
Prototypical Nets~\cite{snell2017prototypical}                & 98.8\%                                                                                                                                     & 99.7\%                                                            &   \\
\cline{1-3}
Meta Nets~\cite{DBLP:journals/corr/MunkhdalaiY17}                        & 98.9\%                                                                                                                                     & -                                                                 &   \\
\cline{1-3}
Snail~\cite{mishra2018a}                            & 99.07 $\pm$ 0.16\%                                                                                                      & \textbf{99.78 $\pm$ 0.09\%}                                          &   \\
\cline{1-3}
MAML~\cite{DBLP:journals/corr/FinnAL17}                             & 98.7 $\pm$ 0.4\%                                                                                                          & \textbf{99.9$\pm$ 0.1\%~}                                             &   \\
\cline{1-3}
\textbf{TAML(Entropy)}           & \textbf{99.23 $\pm$ 0.35\%}                                                                                                                     & 99.71 $\pm$ 0.1\%                                                    &   \\
\cline{1-3}
\textbf{TAML(Theil)}             & \textbf{99.5 $\pm$ 0.3\%}                                                                                                                       & \begin{tabular}[c]{@{}l@{}}\textbf{99.81 $\pm$ 0.1 \%}\\\end{tabular} &   \\
\cline{1-3}
\textbf{TAML(GE(2))}             & \textbf{99.47 $\pm$ 0.25 \%}                                                                                                                     & \begin{tabular}[c]{@{}l@{}}\textbf{99.83 $\pm$ 0.09\%}\\\end{tabular} &   \\
\cline{1-3}
\textbf{TAML(Atkinson)}          & \textbf{99.37 $\pm$ 0.3\%}                                                                                                                      & \textbf{99.77 $\pm$ 0.1\%}                                            &   \\
\cline{1-3}
\textbf{TAML (Gini-Coefficient)} & \textbf{99.3 $\pm$ 0.32\%}                                                                                                                       & 99.70\textit{ }$\pm$ 0.1\%                                             &   \\
\cline{1-3}
\textbf{TAML(GE(0))}             & \textbf{99.33 $\pm$ 0.31\%}                                                                                                                     & \textbf{99.75 $\pm$ 0.09\%}                                             &   \\
\cline{1-3}
\textbf{TAML (VL)}               & \textbf{99.1 $\pm$ 0.36\%}                                                                                                                      & 99.6 $\pm$ 0.1\%                                                        &   \\
\cline{1-3}
\end{tabular}
}%
\vspace{-10pt}
\end{table}

\section{Experiments}\label{sExp}
We report experiment results in this section to evaluate the efficacy of the proposed TAML approaches on a variety of few-shot learning problems on classification and reinforcement learning.
\subsection{Classification}
We use two benchmark datasets Omniglot and MiniImagenet for few-shot classification problem. The Omniglot dataset has 1623 characters from 50 alphabets. Each character has 20 instances which are drawn by different individuals. We randomly select 1200 characters for training and remaining for testing. From 1200 characters, we randomly sample 100 for validation. As proposed in~\cite{DBLP:journals/corr/SantoroBBWL16}, the dataset is augmented with rotations by multiple of 90 degrees.

The Mini-Imagenet dataset was proposed by~\cite{DBLP:journals/corr/VinyalsBLKW16} and it consists of 100 classes from Imagenet dataset. We used the same split proposed by~\cite{Sachin2017} for fair comparison. It involves 64 training classes, 12 validation classes and 20 test classes. We consider 5-way and 20-way classification for both 1-shot and 5-shot.

For $K$-shot $N$-way classification, we first sample $N$ unseen classes from training set and for every $N$ unseen class, we sample $K$ different instances. We follow the same model architecture used by~\cite{DBLP:journals/corr/VinyalsBLKW16}. 
The Omniglot dataset images are downsampled by 28x28 and we use a strided convolutions instead of max-pooling. 
The MiniImagenet images are downsampled to 84x84 and we used 32 filters in the convolutional layers. We also evaluate the proposed approach on non-convolutional neural network. For a fair comparison with MANN~\cite{DBLP:journals/corr/SantoroBBWL16} and MAML~\cite{DBLP:journals/corr/FinnAL17}, we follow the same architecture used by MAML~\cite{DBLP:journals/corr/FinnAL17}. We use Leaky-ReLU as non-linearity instead of ReLU non-linearity.

We train and evaluate the meta-models based on TAML that are unbiased and show they can be adapted to new tasks in few iterations as how they are meta-trained. For Omniglot dataset, we use a batch size of 32 and 16 for 5-way and 20-way classification, respectively. We follow~\cite{DBLP:journals/corr/FinnAL17} for other training settings. For fair comparison with Meta-SGD on 20-way classification, the model was trained with 1 gradient step. For 5-way Mini-Imagenet, we use a batch size of 4 for both 1-shot and 5-shot settings. For 20-way classification on MiniImagenet, the learning rate was set to 0.01 for both 1-shot and 5-shot, and each task is updated using one-gradient step. All the models are trained for 60000 iterations. We use the validation set to tune the hyper-parameter $\lambda$ for both the approaches.

\subsubsection{Results}
We report the results for 5-way Omniglot for both fully connected network and convolutional network. The convolutional network learned by TAML outperforms all the state-of-the-art methods in Table~\ref{table:omni5way}. For 20-way classification, we re-ran the Meta-SGD algorithm with our own training/test splitting for fair comparison since the Meta-SGD is not open-sourced and their training/test split is neither available. The results are reported in the Table~\ref{table:omniconv20way}. It can be shown that TAML outperforms MAML and Meta-SGD for both 1-shot and 5-shot settings.

For MiniImagenet, the proposed TAML approaches outperform the compared ones for 5-way classification problem. The entropy based TAML achieves the best performance compared with inequality-minimization TAML for 5-shot problem. For 20-way setting, we use the reported results from Meta-SGD for both MAML and Meta-SGD. We outperform both MAML and Meta-SGD for both 1-shot and 5-shot settings. It is interesting to note that MAML performs poor compared with matching nets and Meta-learner LSTM when it is trained using one gradient step as reported in Table~\ref{table:miniimagenet}.

\begin{table}
\centering
\caption{Few Shot Classification results on Omniglot dataset for CNN on 20-way setting. For a fair comparison, * denotes re-run results by both meta-learning approaches on the same training/test split used in TAML models. The proposed TAML approaches outperform both MAML and Meta-SGD.}
\label{table:omniconv20way}
\resizebox{0.6\columnwidth}{!}{%
\begin{tabular}{|l|l|l|}
\hline
Methods                            & \multicolumn{2}{l|}{\begin{tabular}[c]{@{}l@{}}~ ~ ~ ~ ~ ~ ~ ~ ~ ~20-way\\~ ~ ~ 1-shot~ ~ ~ ~ ~ ~ ~ ~ ~5-shot~~\end{tabular}}                           \\
\hline
MAML*~\cite{DBLP:journals/corr/FinnAL17}                              & 90.81 $\pm$ 0.5\%                                                                                                                  & 97.49 $\pm$ 0.15\%          \\
\hline
Meta-SGD*~\cite{DBLP:journals/corr/LiZCL17}                          & 93.98 $\pm$ 0.43\%                                                                                                                 & 98.42 $\pm$ 0.11\%          \\
\hline
\textbf{TAML(Entropy + MAML)}      & \textbf{95.62 $\pm$ 0.5\%}                                                                                                         & \textbf{98.64 $\pm$ 0.13\%}  \\
\hline
\textbf{TAML(Theil + Meta-SGD)}    & \textbf{95.15 $\pm$ 0.39\%}                                                                                                        & \textbf{98.56 $\pm$ 0.1\%}   \\
\hline
\textbf{TAML(Atkinson + Meta-SGD)} & \textbf{94.91 $\pm$ 0.42\%}                                                                                                         & \textbf{98.50 $\pm$ 0.1\%}   \\
\hline
\textbf{TAML (VL + Meta-SGD)}      & \textbf{95.12 $\pm$ 0.39\%}                                                                                                        & \textbf{98.58 $\pm$ 0.1\%}   \\
\hline
\textbf{TAML(Theil + MAML)}        & \textbf{92.61 $\pm$ 0.46\%}                                                                                                        & \textbf{98.4 $\pm$ 0.1\%}    \\
\hline
\textbf{TAML(GE(2) + MAML)}        & \textbf{91.78 $\pm$ 0.5\%}                                                                                                          & \textbf{97.93 $\pm$ 0.1\%}    \\
\hline
\textbf{TAML(Atkinson + MAML)}     & \textbf{93.01 $\pm$ 0.47\%}                                                                                                        & \textbf{98.21 $\pm$ 0.1\%}   \\
\hline
\textbf{TAML(GE(0) + MAML)}        & \textbf{92.95 $\pm$ 0.5\%}                                                                                                          & \textbf{98.2 $\pm$ 0.1\%}    \\
\hline
\textbf{TAML (VL + MAML)}          & \textbf{93.38 $\pm$ 0.47\%}                                                                                                        & \textbf{98.54 $\pm$ 0.1\%}   \\
\hline
\end{tabular}
}%
\vspace{-10pt}
\end{table}

\subsection{Reinforcement Learning}
In reinforcement learning, the goal is to learn the optimal policy given fewer trajectories or experiences. A reinforcement learning task $\mathcal{T}{_i}$ is defined as Markov Decision Process that consists of a state space $\mathcal{S}$, an action space $\mathcal{A}$, the reward function $\mathcal{R}$, and state-transition probabilities $q_i(x_{t+1} | x_t,a_t)$ where $a_t$ is the action at time step $t$. In our experiments, we are using the same settings as proposed in \cite{DBLP:journals/corr/FinnAL17}  where we are sampling trajectories using policy $f_\theta$. The loss function used is the negative of the expectation of the sum of the rewards,$
\mathcal{L}_{\mathcal{T}_i} = -\E_{a_t\sim f_\theta,x_t,q_{\mathcal{T}_i}} \left(\sum_{t=1}^T
\mathcal{R}_i(x_t,a_t)\right).$

Experiments were performed using rllab suite~\cite{DBLP:journals/corr/DuanCHSA16}. Vanilla policy gradient~\cite{Williams1992} is used to for inner gradient updates while trust region policy optimizer (TRPO)~\cite{DBLP:journals/corr/SchulmanLMJA15} is used as meta-optimizer. The algorithm is the same as mentioned in algorithm~\ref{ieTaml} with the only difference bing that trajectories were sampled instead of images.

\begin{figure*}
    \subfigure[GE(0)]{
       \includegraphics[width=0.3\textwidth]                   			{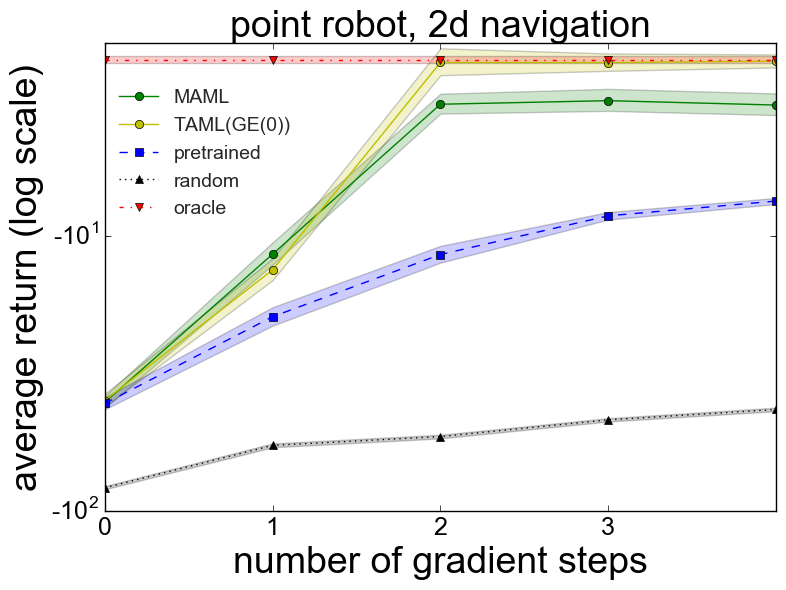}
    }
    \subfigure[Theil]{
       \includegraphics[width=0.3\textwidth]				 	  			{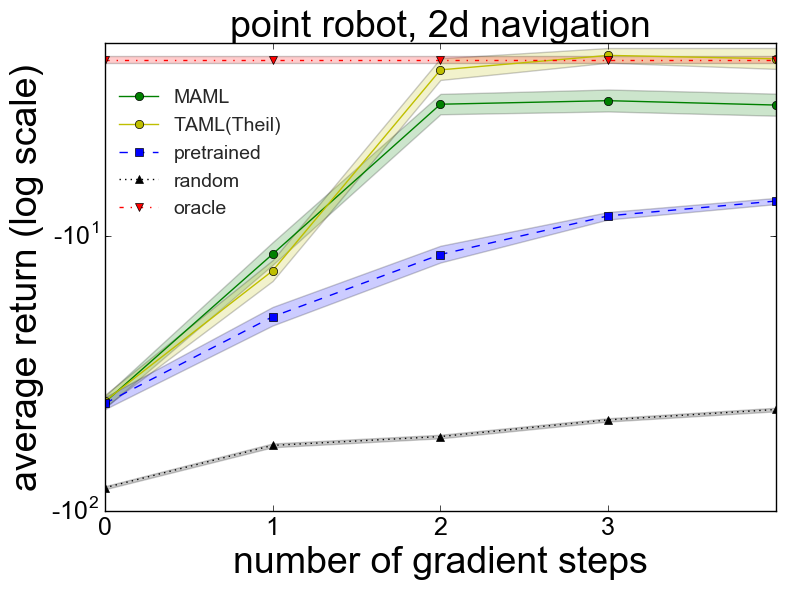}
    }
    \subfigure[GE(2)]{
        \includegraphics[width=0.3\textwidth]
        {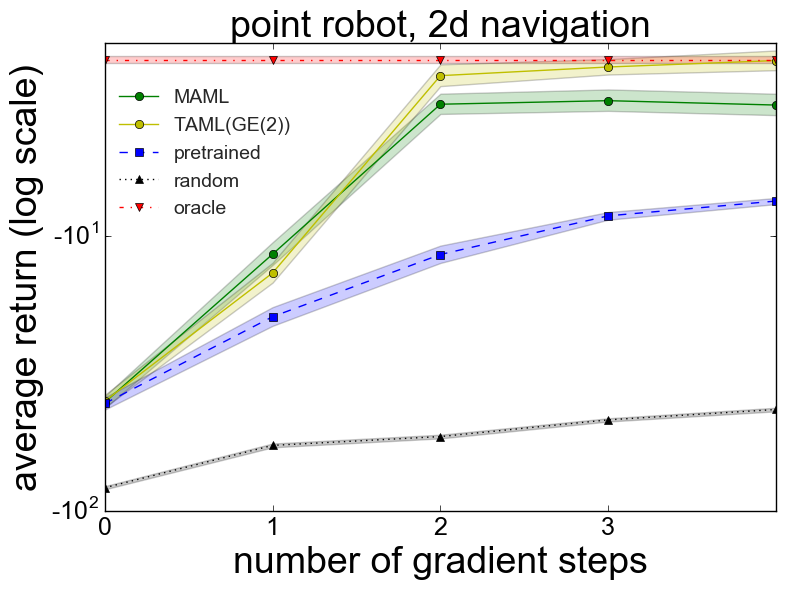}
    }
    \caption{Results on 2D Navigation task. }
    \label{fig:RLexp}
    \vspace{-10pt}
\end{figure*}

For reinforcement learning experiment, we evaluate TAML on a 2D navigation task. The policy network that was used in performing this task is identical to the policy network that was used in \cite{DBLP:journals/corr/FinnAL17} for a fair comparison,  which is a three-layered network using ReLU while setting the step size $\alpha=0.1$. The experiment consists an agent moving in two-dimensional environment and the goal of the agent is to reach the goal state that is randomly sampled from a unit square. For evaluation purposes, we compare the results of TAML with MAML, oracle policy, conventional pre-training and random initialization. Our results have shown that GE(0), Theil, and GE(2) TAML perform on-par with MAML after 2 gradient steps but start to outperform it afterwards as shown in figure~\ref{fig:RLexp}.
\begin{table}
\centering
\caption{Few Shot Classification results on Mini-Imagenet dataset on 5-way and 20-way setting. The results for other methods on 5-way are reported from MAML, and for 20-way, the results are reported from Meta-SGD. TAML approaches outperform MAML on both settings and Meta-SGD on 20-way setting.}
\label{table:miniimagenet}
\resizebox{0.89\columnwidth}{!}{%
\begin{tabular}{|l|l|l|l|l|}
\hline
Methods                         & \multicolumn{2}{l|}{\begin{tabular}[c]{@{}l@{}}~ ~ ~ ~ ~ ~ ~ ~ ~ ~5-way\\~ ~ ~ 1-shot~ ~ ~ ~ ~ ~ ~ ~ ~5-shot~~ \end{tabular}}                          & \multicolumn{2}{l|}{\begin{tabular}[c]{@{}l@{}}~ ~ ~ ~ ~ ~ ~ ~ ~ ~ 20 way\\~ ~ ~ 1-shot~ ~ ~ ~ ~ ~ ~ ~ ~ 5-shot\end{tabular}}                                     \\
\hline
Fine-tune                       & 28.86 $\pm$ 0.54\%                                                                                                                & 49.79 $\pm$ 0.79\%         & -                                                                                                                             & -                                 \\
\hline
Nearest Neighbors               & 41.08 $\pm$ 0.70\%                                                                                                                & 51.04 $\pm$ 0.65\%~        & -                                                                                                                             & -                                 \\
\hline
Matching Nets~\cite{DBLP:journals/corr/VinyalsBLKW16}                   & 43.56 $\pm$ 0.84\%                                                                                                                & 55.31 $\pm$ 0.73\%~        & 17.31 $\pm$ 0.22\%~                                                                                                               & 22.69 $\pm$ 0.20\%                    \\
\hline
Meta-Learn LSTM~\cite{Sachin2017}                 & 43.44 $\pm$ 0.77\%                                                                                                                & 60.60 $\pm$ 0.71\%~        & 16.70 $\pm$ 0.23\%~                                                                                                               & 26.06 $\pm$ 0.25\%                    \\
\hline
MAML (firstorderapprox.)~\cite{DBLP:journals/corr/FinnAL17}        & 48.07 $\pm$ 1.75\%                                                                                                                & 63.15 $\pm$ 0.91\%         & -                                                                                                                             & -                                 \\
\hline
MAML~\cite{DBLP:journals/corr/FinnAL17}                            & 48.70 $\pm$ 1.84\%                                                                                                                & 63.11 $\pm$ 0.92\%~        & 16.49 $\pm$ 0.58\%                                                                                                                & 19.29 $\pm$ 0.29\%                    \\
\hline
Meta-SGD~\cite{DBLP:journals/corr/LiZCL17}                       & 50.47 $\pm$ 1.87\%                                                                                                                & 64.03 $\pm$ 0.94\%~        & 17.56 $\pm$ 0.64\%                                                                                                                & 28.92 $\pm$ 0.35\%                    \\
\hline
\textbf{TAML(Entropy + MAML)}   & \textbf{49.33 $\pm$ 1.8\%}                                                                                                         & \textbf{66.05 $\pm$ 0.85\%} & -                                                                                                                             & -                                 \\
\hline
\textbf{TAML(Theil + MAML)}     & \textbf{49.18 $\pm$ 1.8\%}                                                                                                          & \textbf{65.94 $\pm$ 0.9\%}  & \textbf{18.74 $\pm$ 0.65\%}                                                                                                        & \textbf{25.77 $\pm$ 0.33\%}           \\
\hline
\textbf{TAML(GE(2) + MAML)}     & \textbf{49.13 $\pm$ 1.9\%}                                                                                                          & \textbf{65.18 $\pm$ 0.9\%}  & \textbf{18.22 $\pm$ 0.67\%}                                                                                                        & \textbf{24.89 }$\pm$ \textbf{0.34\%}  \\
\hline
\textbf{TAML(Atkinson + MAML)}  & \textbf{48.93 $\pm$ 1.9\%}                                                                                                          & \textbf{65.24 $\pm$ 0.91\%}  & -                                                                                                                             & -                                 \\
\hline
\textbf{TAML(GE(0) + MAML)}     & \textbf{48.73 $\pm$ 1.8\%}                                                                                                          & \textbf{65.71 $\pm$ 0.9\%}  & \textbf{18.95 $\pm$ 0.68\%}                                                                                                        & \textbf{24.53}$\pm$ \textbf{0.33\%}   \\
\hline
\textbf{TAML (VL + MAML)}       & \textbf{49.4 $\pm$ 1.9\%}                                                                                                          & \textbf{66.0 $\pm$ 0.89\%}  & \textbf{18.13 $\pm$ 0.64\%}                                                                                                        & \textbf{25.33 $\pm$ 0.32\%}           \\
\hline
\textbf{TAML(GE(0) + Meta-SGD)} & -                                                                                                                             & -                      & \textbf{18.61 $\pm$ 0.64\%}                                                                                                        & \textbf{29.75}$\pm$ \textbf{0.34\%}   \\
\hline
\textbf{TAML (VL + Meta-SGD)}   & -                                                                                                                             & -                      & \textbf{18.59 $\pm$ 0.65\%}                                                                                                        & \textbf{29.81 }$\pm$ \textbf{0.35\%}  \\
\hline
\end{tabular}
}%
\vspace{-10pt}
\end{table}

\section{Conclusion}
In this paper, we proposed a novel paradigm of Task-Agnostic Meta-Learning (TAML) algorithms to train a meta-learner unbiased towards a variety of tasks before its initial model is adapted to unseen tasks. Both an entropy-based TAML and a general inequality-minimization TAML applicable to more ubiquitous scenarios are presented.  We argue that the meta-learner with unbiased task-agnostic prior could be more generalizable to handle new tasks compared with the conventional meta-learning algorithms.  The experiment results also demonstrate the TAML could consistently outperform existing meta-learning algorithms on both few-shot classification and reinforcement learning tasks.


\small{
\bibliographystyle{unsrt}
\bibliography{taml}
}
\end{document}